\pdfoutput=1
\documentclass[11pt]{article}

\usepackage{authblk}
\usepackage{emnlp2021}

\usepackage{times}
\usepackage{latexsym}

\usepackage[T1]{fontenc}
\usepackage[utf8]{inputenc}

\usepackage{microtype}

\usepackage{makecell}

\usepackage{url}
\usepackage{tabularx}

\usepackage{float}
\usepackage{graphicx}
\usepackage[caption=false]{subfig}

\usepackage{multirow}
\usepackage{enumitem}

\usepackage{amssymb}
\newcommand{\infounit}{$\mathbb{IU}$}
\newcommand{\sect}{\S}
\newcommand{\super}{SuperPAL}
\newcommand{\RA}{\textsc{Rouge}}

\usepackage{color,soul}

\usepackage{todonotes}

\title{Summary-Source Proposition-level Alignment:\\ Task, Datasets and Supervised Baseline}
\author[1]{\bf Ori Ernst}
\author[1]{\bf Ori Shapira}
\author[2]{\bf Ramakanth Pasunuru}
\author[1]{\bf Michael Lepioshkin}
\author[1]{\\ \bf Jacob Goldberger}
\author[2]{\bf Mohit Bansal}
\author[1]{\bf Ido Dagan}
{
\makeatletter
\renewcommand\AB@affilsepx{~~~~~~ \protect\Affilfont} \makeatother
\affil[1]{Bar-Ilan University}
\affil[2]{UNC Chapel Hill}
}
\affil[  ]{} % skip line of affiliations
\affil[  ]{\tt \{oriern, obspp18, mishamikel\}@gmail.com}
\affil[  ]{\tt \{ram, mbansal\}@cs.unc.edu}
\affil[  ]{\tt \{jacob.goldberger@, dagan@cs.\}biu.ac.il}

\date{}

\begin{document}
\maketitle

\begin{abstract}
Aligning sentences in a reference summary with their counterparts in source documents was shown as a useful auxiliary summarization task, notably for generating training data for salience detection. Despite its assessed utility, the alignment step was mostly approached with heuristic unsupervised methods, typically ROUGE-based, and was never independently optimized or evaluated. In this paper, we propose establishing summary-source alignment as an explicit task, while introducing two major novelties: (1) applying it at the more accurate proposition span level, and (2) approaching it as a supervised classification task. To that end, we created a novel training dataset for proposition-level alignment, derived automatically from available summarization evaluation data. In addition, we crowdsourced dev and test datasets, enabling model development and proper evaluation. Utilizing these data, we present a supervised proposition alignment baseline model, showing improved alignment-quality over the unsupervised approach.
\end{abstract}

\section{Introduction}
\label{sec_intro}

Text summarization aims to extract the salient information out of a single document or a set of topically-related documents, and to generate a coherent summary.
Inherently, summarization needs to address, either explicitly or implicitly, several embedded subtasks, such as salience detection, redundancy removal, and text generation.

\begin{figure*}[t!]
    \centering
    \resizebox{\linewidth}{!}{
    \includegraphics{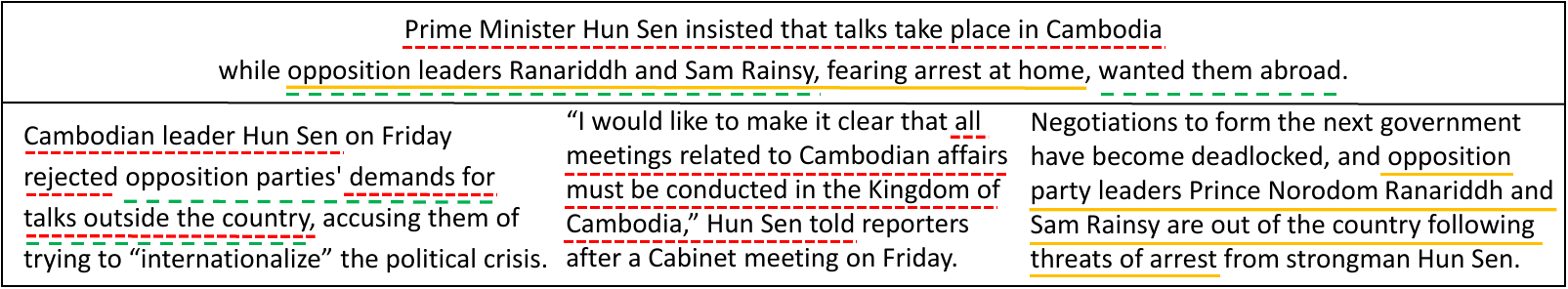}}
    \caption{Aligning \infounit s between a summary sentence (top) and sentences from documents (bottom). For fuller example see Appendix \ref{sec_full_alignment_example}.}
    \label{fig_alignmentExample}
\end{figure*}

Attempting to cope with this encompassing challenge, summarization research often involves developing models for specific summarization subtasks, utilized either as system components or for auxiliary purposes such as training data creation.
In this paper, we draw attention to the particular auxiliary task of \textit{summary-source alignment}, which was utilized as a supporting data generation step in the summarization literature, for both single and multi document summarization. Given a gold summarization dataset, the task aligns information pieces in a reference summary with corresponding information in the source documents. 
Such alignments were generated automatically over large summarization datasets, most typically at the full sentence level. Then, (noisy) training data for certain summarization subtasks was automatically derived from these alignments (\sect \ref{sec_relatedWork}), notably for salience detection \citep{gehrmann2018bottomup, chen2018fastAbsSumm, Lebanoff2019ScoringSS}, but also for redundancy recognition \citep{Cho2019ImprovingTS} and text rephrasing and fusion  \citep{zhang2018neuralExtractiveSumm, Lebanoff2019ScoringSS}. 
Even though the quality of the subsequent trained models relies on alignment quality, the intermediate alignment methods were neither optimized nor evaluated explicitly, making them difficult to compare and improve.

In this paper, we establish summary-source alignment as a stand-alone auxiliary task, with corresponding novel datasets, model, and intrinsic evaluation.
As a major contribution, aiming to yield more precise alignments, we propose aligning source-summary information at the more fine-grained level of \textit{propositions}, rather than at the common sentence level.
Specifically, we align information at the level of individual proposition spans, termed \textit{information units} (\infounit s), as exemplified in Fig. \ref{fig_alignmentExample}. This level provides much tighter alignments compared to the sentence level used in prior work, since aligned full sentences would typically include both matching propositions as well as non-matching ones. An illustration for the tighter alignment at the proposition-level, over random examples, is shown in Table \ref{tab_ROUGE_prop_sent}, where full sentences usually include additional non-matching content. Table \ref{tab_ROUGE_prop_sent} also demonstrates that the ROUGE-L score between matching propositions tends to be much higher than between sentences, unless the semantic matching is abstractive (rather than lexical), where ROUGE is not informative.

To support research on the advocated proposition-level alignment, we first developed an elaborate crowdsourcing methodology and created high-quality development and test datasets (\sect \ref{sec_dataCollection}). Next, we automatically derive a larger-scale training dataset consisting of 23K alignment instances from available Multi-Document Summarization (MDS) evaluation data, available as reliable Pyrmaid annotations \cite{nenkova2004pyramid} (\sect \ref{sec_pyrData}). This data is utilized to train a supervised alignment baseline model (\sect \ref{sec_spanExperiments}), which outperforms traditional unsupervised alignment approaches.\footnote{All corresponding datasets and code are publicly available at \href{https://github.com/oriern/SuperPAL}{https://github.com/oriern/SuperPAL}.} 
Moreover, thanks to this novel training dataset, we show (\sect \ref{subsec_span_results}) that our baseline aligner is capable of producing ``abstractive'' alignments, where there is almost no lexical overlap, while traditional aligners fail to do so. We further show intrinsically that our proposition-level aligner extracts better salient sentences than common sentence-level aligners. 
Notably, while our datasets are derived from MDS sources, the data and model are applicable also for alignments over Single Document Summarization data. 
In concluding discussion, we suggest using our dataset suite to further develop improved proposition-level aligners, which in turn may trigger appealing research on proposition-based summarization methods.

\begin{table*}[t]
    \resizebox{\linewidth}{!}{
    \begin{tabular}{l|l|c}
    \multicolumn{1}{c|}{\bfseries {Summary Sentence}} & \multicolumn{1}{c|}{\bfseries {Document Sentence}} & \multicolumn{1}{c}{\bfseries {R-L}} \\ \hline

     The BBC reports 56-year-old \textcolor{red}{\textit{Allan Matthews}} pleaded guilty&   It is the crown's case that \textcolor{red}{\textit{Matthews is not qualified}} & \multirow{3}{*}{17.85 / \textcolor{red}{\textit{28.57}}} \\
        Wednesday to removing another man's left hand at an & \textcolor{red}{\textit{or authorised to perform such a procedure, and is}}  &\\
       Australian motel despite \textcolor{red}{\textit{not being qualified to practice medicine.}} & \textcolor{red}{\textit{not a qualified or registered medical practitioner.}}  &\\ \hline

        \textcolor{red}{\textit{The lawsuit, which also alleges a hostile work environment}} &  \textcolor{red}{\textit{Campbell alleges that events like those construe a}}  & \multirow{5}{*}{25.00 / \textcolor{red}{\textit{50.00}}} \\
          and retaliation, claims the sexist culture at Magic Leap created a& \textcolor{red}{\textit{hostile working environment,}} and is asking for & \\
            "dysfunctional" workplace and is part of the reason the company& punitive damages from Magic Leap. & \\
              has yet to actually release a product...&   &\\ \hline

      \textcolor{red}{\textit{ A U.S. House resolution criticized Hun Sen's regime}}  &   Cambodia's ruling party responded Tuesday to \textcolor{red}{\textit{criticisms}}& \multirow{3}{*}{20.40/ \textcolor{red}{\textit{21.05}}} \\
        while the opposition tried to cut off his access to loans. & \textcolor{red}{\textit{of its leader in the U.S. Congress}} with a lengthy  & \\
      &  defense of strongman Hun Sen's human rights record.  &\\ \hline

      \textcolor{red}{\textit{Ecevit}}, a former prime minister, \textcolor{red}{\textit{was asked to form a }}&   \textcolor{red}{\textit{Ecevit must now try to build a government}} that includes & \multirow{2}{*}{25.80/ \textcolor{red}{\textit{50.00}}} \\
         \textcolor{red}{\textit{new government.}} & their center-right parties but not them as individuals.   &\\

    \end{tabular}}
    \caption{ROUGE-L score between \textcolor{red}{\textit{aligned \infounit s}} and their corresponding sentences.}
    \label{tab_ROUGE_prop_sent}
\end{table*}

\section{Background and Related Work}
\label{sec_relatedWork}
As mentioned above, several methods leveraged automatically generated reference-source sentence alignments, to derive (noisy) training sets for summarization subtasks \citep{zhang2018neuralExtractiveSumm, Cho2019ImprovingTS}.
For example, training datasets for sentence salience detection were derived from reference-source alignments, by marking as salient those source sentences that were aligned with a summary sentence (e.g. \citet{chen2018fastAbsSumm}).
As another example, \citet{Lebanoff2019ScoringSS} leveraged alignments to create a sentence fusion dataset: the input for each fusion instance consists of a pair of source sentences that are aligned to the same summary sentence, while the aligned summary sentence is regarded as the output fused sentence.

The underlying sentence alignments, from which the above training datasets were derived, were extracted automatically from large summarization datasets. Alignments were detected using unsupervised sentence similarity measures, typically based on ROUGE score \citep{lin2004rouge} (see \sect \ref{sec_sentExperiments} for more details).
Typically, models trained over the alignment-based datasets were evaluated only on the final summary. Yet, alignment quality, which determines the quality of the utilized training datasets, was never optimized or assessed explicitly, as we do in this paper.

Notably, alignment of matching pieces of information provided the basis for the prominent Pyramid method for summarization evaluation \citep{nenkova2004pyramid}, capturing information overlap between system summaries and reference summaries.
Alignments were performed at the level of individual propositions, termed Summary Content Units (SCUs) (similar to the information units marked in Fig.~\ref{fig_alignmentExample}).
Matching information at the proposition level was favored over the more coarse sentence level, since a system summary sentence may include some propositions that match the reference and some that don't.
Later works attempted to automate the Pyramid procedure, using Open IE \citep{Yang2016PEAKPE,peyrard-eckle-kohler-2017-supervised}  or Elementary Discourse Units (EDU) \citep{hirao-etal-2018-automatic}, to extract proposition-level units. As propositional units (like SCUs) may share their arguments (such as in conjunctions and other constructions,  e.g. “The boy went home and ate dinner”), and may be discontiguous (“The boy...ate dinner”), Open IE output, which satisfies these requirements, is best suitable for extracting such units (while EDU format does not satisfy these properties).
Inspired by these works, we align proposition spans (\sect \ref{sec_dataCollection}), extracted automatically using Open IE (\sect \ref{sec_spanExperiments}).

\section{Dev and Test Alignment Datasets}
\label{sec_dataCollection}
This section presents the manually-annotated development and test datasets for reference-source alignments, including their structure (\sect \ref{subsec_data_structure}), source data (\sect \ref{subsec_dataCollection_baseData}) and annotation process (\sect \ref{subsec_annotation}). 
These datasets allow direct tuning and evaluation of alignment algorithms, lacking in prior work (\sect \ref{sec_intro}).

\subsection{Dataset Structure}
\label{subsec_data_structure}

In the typical MDS setting, a summarization instance consists of a set of topically-related documents, often termed a \textit{topic}, and corresponding gold reference summary(ies) \cite{nist2014duc, fabbri2019multinews}.
For such an instance, we collect all alignments
between each proposition span in the reference summary and the corresponding propositions, conveying the same information, in the source documents. 
We choose the proposition-span level, termed \textit{information unit} (\infounit), as the basis for alignment following the rationale of similar SCU-level alignments in the well-established Pyramid evaluation method (\sect \ref{sec_relatedWork}).
To facilitate crowdsourcing, we adapt a somewhat looser definition for our \infounit s (Task 1 below). 
Figure \ref{fig_alignmentExample} illustrates some \infounit~spans and their alignments.

\subsection{Source Data}
\label{subsec_dataCollection_baseData}
We leverage three MDS datasets to create our alignment dataset: DUC 2004, DUC 2007 \citep{nist2014duc}, and the recent Multi-News \citep{fabbri2019multinews}. 
We sampled 21 topics: 9 topics (4 dev, 5 test) from MultiNews, each with 3-4 documents and 1 reference summary, and 6 topics (3 dev, 3 test) from each of the DUC datasets, sampling 7 documents and 1 reference summary per topic.

\begin{figure*}[t!]
    \centering
    \resizebox{\linewidth}{!}{
    \includegraphics{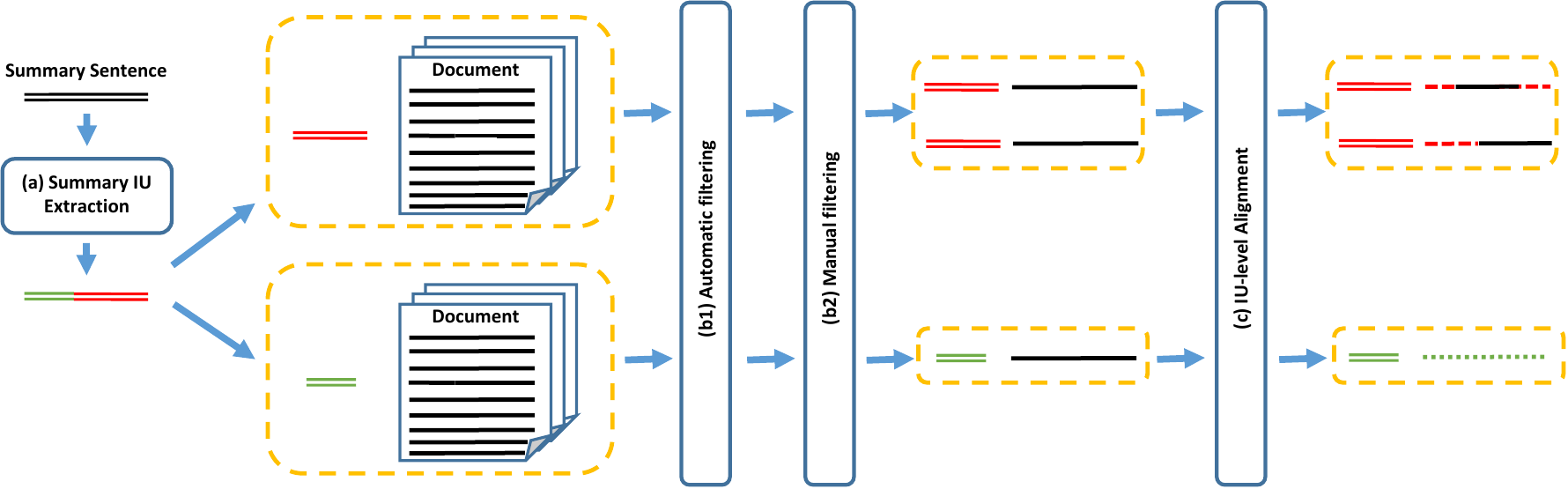}}
    \caption{Propositional alignment annotation process: A summary sentence is divided into \infounit s  ((a), \textbf{Task 1}). Then, each \infounit~is paired with all document sentences, yielding sets of candidate pairs. These sets are filtered automatically, and then manually, to reduce annotation complexity ((b1) \& (b2), \textbf{Task 2}). Finally, workers mark the aligned propositional \infounit s in the remaining candidate pairs ((c), \textbf{Task 3}).}
    \label{fig_annotation_process}
\end{figure*}

\subsection{Annotation}
\label{subsec_annotation}
Our annotation process relied mostly on crowdsourcing, while involving limited expert effort to obtain high quality dev and test sets.
The crowdsourcing process was divided into three tasks, conducted on Amazon Mechanical Turk with total cost of \$3,163.
We required workers from native English speaking countries, with $>98\%$ approval rate and $>500$ approved tasks. 

To estimate the recall of the crowdsourced annotations in each task, we tested them, as reported below, against meticulous expert annotation of 6 topics, conducted by one of the authors (using the DUCView annotation tool \citep{sigelman2006ducview}).
Further, to assure alignment precision, all obtained crowdsourced alignments were filtered or corrected by one of the authors, thus ensuring expert-level precision for our dataset.

To evaluate one alignment against another (such as  crowdsourced against expert), we first need to determine whether a span identified in one alignment should be considered as matching a span in the other. Since determining exact proposition boundaries may be somewhat subjective, we follow an ``intersection over union" soft matching approach, often used in similar cases (e.g. \cite{roit-etal-2020-controlled}).
In our case, we apply Jaccard similarity, measuring intersection-over-union between the two spans' character-level position indices with respect to the beginning of their sentence. For two span annotations $span_{A}, span_{B}$ in  sentence $sent$, the corresponding character position indices sets are:
$A=\{i \mid sent[i] \in span_{A}\}$, $B=\{i \mid sent[i] \in span_{B}\}$, respectively.
The Jaccard score is then defined as:
\begin{equation}
\label{equation_scoreP}
    Jaccard=\frac{A\cap B}{A\cup B}
    \label{eq_jac}
\end{equation}
As explained below, we tuned a threshold $t$ for the Jaccard similarity, above which two spans are considered as matching. 

Next we describe the three tasks of our annotation pipeline. The full process is illustrated in Figure \ref{fig_annotation_process}, and full annotation guidelines can be found in Appendix \ref{sec_croud_guidelines}.

\textbf{\underline{Task 1:} Summary \infounit~extraction.}\hspace{0.5cm} In this task, workers are asked to mark all \infounit~spans within the reference summary sentences. The task instructions, accompanied with illustrating examples, define an \infounit~as a standalone fact, covering the (possibly non-consecutive) span of a predicate and all its arguments, as exemplified in Fig. \ref{fig_alignmentExample}.\footnote{The annotation instructions for all tasks are embedded in the annotation interface, included in our Github release.}
To calculate the recall of the crowdsourced annotation, we consider an expert \infounit~as being matched if its Jaccard score with at least one crowdsourced \infounit~is beyond a threshold $t$. Through manual examination, we found that $t=0.25$ closely approximates appropriate matches, yielding almost perfect coverage of the expert \infounit s. We adopt this tuned threshold for evaluating the output of the next tasks as well. Overall, we collected 203 dev and 238 test \infounit s.

\textbf{\underline{Task 2:} Filtering unlikely candidate alignments.}\hspace{0.5cm}
In Tasks 2 and 3, we find, for each summary \infounit, all its aligned \infounit s in the source documents of the topic. In Task 2, for each reference summary \infounit, we filter source document sentences that are  unlikely to include an aligned \infounit, keeping all remaining sentences as candidates for alignment. Then, in Task 3, we ask annotators to verify  whether alignment indeed holds, and if so, annotate it at the \infounit-span level. Task 2 thus reduces the burden on the actual \infounit~alignment annotation in Task 3, filtering the vast amount of irrelevant source sentences for each summary \infounit.

The Task 2 filtering process consists of two steps, automatic and crowdsourced.
First, we automatically filter candidate pairs of a summary \infounit~and a document sentence that do not satisfy a certain similarity criterion, which is is composed of several similarity scores:
the BERT-based similarity measure BERTscore \citep{zhang2019bertscore}, an entailment score based on RoBERTa fine-tuned on MNLI \citep{liu2019roberta}, and ROUGE-1 precision \citep{lin2004rouge}.\footnote{For BERT and ROUGE scores, the summary \infounit~ is considered as the candidate and the document sentence as the reference; for entailment score, the summary \infounit~is the hypothesis and the document sentence is the premise.} Next, we filter the remaining pairs via crowdsourcing. Given a summary \infounit, bolded within its sentence, and a candidate document sentence, the workers determine whether
the source sentence contains an \infounit~that should be aligned with the given summary \infounit. 
Overall, 94\% of the candidate summary \infounit~and sentence pairs were filtered in Task 2, while yielding a recall of $83\%$ relative to the expert alignments.

\textbf{\underline{Task 3:} \infounit-level alignment.}\hspace{0.5cm}In this task, annotators are given a summary \infounit, highlighted in bold within its sentence, and a candidate source sentence, which was judged in Task 2 as being aligned with the summary \infounit. The worker should then mark the aligned \infounit~span in the source sentence, or state that no alignment resides. As mentioned earlier, these alignment annotations underwent a final expert review, eventually yielding 345 development and 312 test reference-source \infounit~alignments (with expert-level precision).

We then measured the recall of the alignments obtained in Tasks 2 and 3 against the expert annotation (of 6 topics, mentioned above). An expert alignment was considered as covered if it matched at least one crowdsourced alignment, where the Jaccard match score was above $t=0.25$ for both the summary and source sides, yielding 76\% recall. The average Jaccard similarity of the matched alignments in our data is about 0.85, indicating that most alignments were off by only one or two words relative to expert annotation.
Overall, our (dev and test) crowdsourced dataset is the first to provide effective means for tuning and comparing alternative alignment models , as shown in Section \ref{sec_spanExperiments}.

\section{Pyramid-Based Training Dataset}
\label{sec_pyrData}
To obtain larger amounts of alignments for training supervised models, 
we derive them automatically from existing MDS evaluation data. While these alignments are less exhaustive, compared to the manual processes in \sect \ref{sec_dataCollection}, they are of fairly high quality and proved useful for model training.

We follow Copeck et al.'s line of work \citep{Copeck2005LeveragingP, Copeck2006LeveragingD, Copeck2007CatchWY, Copeck2008UpdateSU}, which established a sentence alignment dataset based on the Pyramid evaluation method \cite{nenkova2004pyramid}, applied for the DUC 2005-2007 and TAC 2008 summarization benchmarks. As many of the systems participating in these benchmarks were extractive, system summary sentences directly link to document sentences. These links, along with the Pyramid's expert mapping between reference summary spans and system summary spans, enabled Copeck et al. to transitively align reference summary sentences to document sentences, yielding a \textit{sentence-level} alignment dataset.

We take this process one step further, aligning in a similar manner \infounit~\textit{spans} in the reference summaries (rather than full sentences) to corresponding \infounit~\textit{spans} in documents. This is possible thanks to the Pyramid's annotated mappings, which link spans between reference and system summaries. Such derived alignments are not exhaustive, since the Pyramid-based alignments cover only sentences included in the evaluated system summaries. Nevertheless, this provides a novel large-scale alignment dataset at the proposition-level, consisting of 18,505 alignments, sufficient for training neural alignment models.

The alignments obtained thus far involve spans that were detected manually by the Pyramid annotators. Yet, our alignment model (Section \ref{sec_spanExperiments}) needs to align \infounit s that were extracted automatically, a step that we implemented using OpenIE (OIE) \citep{Stanovsky2018SupervisedOI}. In order to train our alignment model appropriately, we create an additional version of the training set, in which the aligned \infounit s were detected automatically by our OIE-based extraction (Section \ref{sec_spanExperiments}). To that end, we first apply the \infounit~extractor over the summaries and source documents. Then, we consider a candidate pair of (OIE-based) summary \infounit~and source \infounit~as matching if each of these \infounit s matches the corresponding \infounit s in a Pyramid-based alignment, with a Jaccard similarity score above the $t=0.25$ threshold (based on Section \ref{sec_dataCollection}). This process yielded 23,492 OIE-based alignments, as some of the original spans matched several OIE spans, while others did not match any.

Next, we add negative instances to our dataset. First, we include challenging negative instances by selecting all non-aligning source-summary OIE combinations that have BERTscore \citep{zhang2019bertscore} above 0.89. In addition 
we include naturally-distributed negative instances, 
by taking a summary \infounit~that has an alignment in a certain document and  combining it with all non-aligned candidate \infounit s in that document. 
Since our training dataset is based on alignment to extractive system summaries, we consider in this process only document sentences that are included in these system summaries. Overall, we sampled 219,772 negative examples, where 22\% of them were selected for high BERTscore.

We estimated our data quality by manually counting errors in random samples, including 80 positive examples and 80 negative examples of each of the two negative example types. The respective error rates were 11\% in the positive sample and 5\% in each of the negative samples, suggesting the high-quality of our automatically-derived data. Its actual effectiveness for model training is assessed next. A few samples from the data are presented in Table \ref{tab_train_example}.

\renewcommand\tabularxcolumn[1]{m{#1}}
\newcolumntype{?}{!{\vrule width 2pt}}

\begin{table*}[t]
    \footnotesize
    
    \resizebox{\linewidth}{!}{
    \begin{tabularx}{\textwidth}{cX|X?X|X}%{lX}
    % \begin{tabular}{l|l|l|l}
   &\centering{\bfseries {Orig Summary Span}} & \centering{\bfseries {OIE Summary Span}} &
     \centering{\bfseries {Orig Document Span}} &
      \multicolumn{1}{c}{\bfseries {OIE Document Span}} \\ \hline

    (a)&
    \rule{0pt}{3ex}
    European Union (EU) nations agreed that a single currency (the Euro) will go into effect on January 1, 1999 &
    that a single currency ... Euro) will go into effect on January 1, 1999. &
    the 1999 introduction of the single European currency, the euro &
    The mass printing of the banknotes of the single European currency ... euro ... would be started at the beginning of 1999  \\ \hline

    (b)&
    president's line-item veto...of 38 military construction projects &
    \rule{0pt}{3ex}
    In October 1997 Congress overrode the president's line-item veto against 36 of 38 military construction projects. &
    U.S. President Bill Clinton used his line-item veto...to strike out...projects...from a military construction bill &
    his line-item veto power to strike out 38 projects worth 287 million U.S. dollars \\ \hline

    (c)&
    \centering N/A &
    
    500 anti-government activists surrendered in March 1999. &
    \centering N/A &\rule{0pt}{3ex}
    a total of 15 anti-government ethnic armed groups have made peace with the government \\ \hline
    
    (d)&
    \centering N/A &
    
    the suit filed by the recording industry. &
    \centering N/A &\rule{0pt}{3ex}
    Metallica...alleging that Napster's software encouraged users to freely trade the band's songs without permission. \\ \hline

    (e)&
    \centering N/A &
    
    US tobacco companies appear capable of sustaining strong momentum. &
    \centering N/A &\rule{0pt}{3ex}
    yesterday reported a more modest 10.9 per cent advance in net earnings \\ \hline

    % \end{tabular}}
     \end{tabularx}}
    \caption{Examples from alignment training data. (a) and (b) are positive examples, where original aligned spans have been extracted manually by Pyramid annotators. For our data, we used only OpenIE spans with high overlap with the original ones. (c) and (d) are challenging negative examples with high BERTscore, while (e) is taken from the naturally distributed negative sample. Negative samples were extracted in OpenIE format directly, and do not have original annotated spans (N/A).}
    \label{tab_train_example}
\end{table*}

\section{\infounit~Alignment Methods and Evaluation}
\label{sec_spanExperiments}

This section describes two baseline methods (\sect \ref{subsec_lex_model}, \sect \ref{subsec_ensemble_model}) and our proposed supervised method (\sect \ref{subsec_supervised_model}) for summary-source \infounit-level alignment.
We compare their performance (\sect \ref{subsec_span_results})
against our manually-created test set, showing the substantial advantage of the supervised aligner, trained over
our automatically-derived training dataset (\sect \ref{sec_pyrData}).

The presented alignment methods require a preliminary step that extracts candidate \infounit~spans to be aligned. As mentioned earlier, we found Open IE (OIE) \citep{Stanovsky2018SupervisedOI} suitable for this purpose, simply collecting as a (possibly non-consecutive) \infounit~span the union of a predicate and its arguments in an OIE extraction.
Using OIE loses around 10\% of the correct (gold) alignments due to missed \infounit~extractions, limiting the automatic alignment recall to an upper-bound of 90\%.

\subsection{ROUGE-based Lexical Model}
\label{subsec_lex_model}
As described in \sect \ref{sec_relatedWork}, the most common alignment approach, previously applied at the sentence level, is based on ROUGE lexical similarity. Typically, a reference summary sentence is aligned with one or two source sentences which are most similar to it.
We adjust this approach to the \infounit~level, denoted $\RA_{IU}$. Accordingly,
each summary \infounit~is matched with the $k$ document \infounit s of highest ROUGE similarity, choosing $k=2$, which worked best on the dev set.\footnote{Our ROUGE similarity is an average of recall R-1, R-2, and R-L, where the summary-\infounit~is considered the reference. We also experimented with setting a threshold over the similarity score, but the common top-$k$ approach worked best.}

\subsection{Semantic Similarity Ensemble Model}
\label{subsec_ensemble_model}
As a distantly supervised approach for matching semantically-equivalent \infounit s, we developed and tuned an ensemble of various existing semantic matching models, denoted ``Sim-Ensemble''. Specifically, we ended up with a two-stage approach, where we first align a summary \infounit~with the three source sentences most similar to it. 
The similarity score at this stage was a tuned combination of ROUGE, RoBERTa-MNLI \citep{liu2019roberta} and BERTscore \citep{zhang2019bertscore}.
Then, to find aligned spans, we applied a word aligner \citep{sultan2014back} to align words between a document sentence and the summary \infounit, and aligned the consecutive text spans between the first and last aligned words on each side (filtering pairs with too few word alignments).

\begin{table}[t]
\centering
    \resizebox{\linewidth}{!}{
    \begin{tabular}{cc|c|c|c|c|c}
    &    \textbf{Ref.} & $R_{0.25}$ & $P_{0.25}$ & $F_{1}$ & $Cov_{0.25}$ & $F_{1,cov}$  \\ \hline 
      \multirow{2}{*} {$\RA_{IU}$} 
    &    Dev Set  & 33.91 & 37.76 & 35.73 & 50.26 & 43.12  \\
    &    Test Set & 28.85 & 29.97 & 29.40 & 47.34 & 36.71  \\\hline 
      \multirow{2}{*} {Sim-Ensemble} 
    &    Dev Set  & 43.48 & 41.01 & 42.21 & 59.79 & 48.65 \\
    &    Test Set & 37.18 & 34.77 & 35.93 & 52.66 & 41.89 \\\hline 
      \multirow{2}{*} {\super} 
    &    Dev Set  & 47.83 & 66.29 & \textbf{55.56} & 55.56 & \textbf{60.45}\\
    &    Test Set & 43.59 & 65.85 & \textbf{52.46} & 54.44 & \textbf{59.60} \\\hline 
    
    \end{tabular}}
    \caption{Automatic aligners scores}
    \label{tab_baseline}
    
\end{table}

\subsection{Supervised Model}
\label{subsec_supervised_model}
This model is a binary classifier, deciding whether two given \infounit s align.
We follow the standard usage of RoBERTa for paraphrasing tasks \citep{liu2019roberta}. Specifically, we take a RoBERTa encoder fine-tuned on MNLI and augment it with a new binary classification layer.
This model is then further fine-tuned  with our Pyramid-based dataset (\sect \ref{sec_pyrData}). We denote this model as ``\super'', for Supervised Propositional ALignment.

\subsection{\infounit-level Results}
\label{subsec_span_results}
The alignment methods are evaluated using the same character-level Jaccard similarity, which was used for evaluating the crowdsourcing annotations (Eq. \ref{eq_jac}). Here, we define recall ($R_{0.25}$), precision ($P_{0.25}$) and F-1 ($F_1$), where a predicted and a gold alignment are considered matching if they yield a Jaccard score surpassing the threshold $t=0.25$ for the \infounit~pairs in both summary and document sides. In addition, since alignment may be utilized eventually to compose a summary, the requirement of finding \textit{all} aligned document \infounit s for each summary \infounit, as measured by $R_{0.25}$, might be superfluous. To that end, a \textit{Coverage} ($Cov_{0.25}$) measure was added, measuring the proportion of summary \infounit s covered by at least one aligned pair. Respectively, an $F_{1,cov}$ balances between $Prec_{0.25}$ and $Cov_{0.25}$. The methods are evaluated against our gold dev and test datasets, as in Table \ref{tab_baseline}.

As shown, the \super~model substantially outperforms the other two baselines. The lexical ROUGE-based baseline, typical of prior work (which was not evaluated intrinsically), performs worst.
The Sim-Ensemble model, which was trained on generic NLI and text similarity data, surpasses the unsupervised ROUGE-based model by 6 $F_1$ points. Yet, it scores 16 $F_1$ points lower than the supervised model, which was trained for the \infounit~alignment task using our new training data. Finally, the average Jaccard score for matching predicted and gold \infounit s for  \super~is 0.67, indicating a high match with gold annotation.

It is further illuminating to examine the results of $\RA_{IU}$ and \super~by a breakdown over the DUC and MultiNews parts of our datasets, shown in Table \ref{tab_spanDUC_MN}.
Clearly, while DUC is generally more challenging, the performance of \super~degrades more gracefully relative to MultiNews, while the $\RA_{IU}$ performance on DUC collapses drastically. 
This might be caused by the more abstractive nature of DUC relative to MultiNews \cite{fabbri2019multinews}, suggesting that due to its lexical nature 
$\RA_{IU}$ is inadequate to identify abstractive alignments with low lexical overlap.
In addition, a manual inspection suggested that $\RA_{IU}$ tends to be misled by non-aligned \infounit s that exhibit high lexical overlap, as exemplified in Table \ref{tab_RougeExmp}. On the other hand,  \super~is more capable in identifying lexically-dissimilar paraphrases, as exemplified in Table \ref{tab_supervisedExmp}. This behavior may be attributed to the large abstractive dataset on which our model was trained, along with the use of a pre-trained contextualized embedding model, which is known to capture semantic similarities. Overall, the much larger gap between the two models for the more abstractive DUC suggests the appeal of \super~for abstractive semantic matching.

To enable development of new proposition-based summarization methods using our alignments (as explained in \sect \ref{sec_conclusion}), we release 
our proposition-based alignments, produced by our supervised aligner,
for the MultiNews MDS dataset \citep{fabbri2019multinews} and for the CNN/DailyMail Single Document Summarization dataset \citep{Hermann2015TeachingMT}.

\begin{table}[t]
    \resizebox{\linewidth}{!}{
    \begin{tabular}{cl|c|c|c|c|c}
    &\textbf{Method} & $R_{0.25}$ & $P_{0.25}$ & $F_1$ & $Cov_{0.25}$ & $F_{1,cov}$ \\ \hline

     \multirow{2}{*}{DUC}&    $\RA_{IU}$  & 29.25 & 23.71 & 26.19 & 44.05 & 30.82	 \\

         &\super~& 36.73 & 67.59 & \textbf{47.59} & 40.48 &  \textbf{50.63}   \\ \hline
         
     \multirow{2}{*}{MN}&    $\RA_{IU}$  & 36.97 & 48.43	& 41.93 & 55.68 & 51.80	 \\

         &\super~& 49.09 &  64.2 & \textbf{55.63} &60.23 &  \textbf{62.15}   \\
    \end{tabular}}
    \caption{Span alignment scores of $\RA_{IU}$ and \super~aligners, on the DUC and MultiNews datasets. Examined on the dev \& test sets together.}
    \label{tab_spanDUC_MN}
\end{table}

\begin{table}[t]
    \resizebox{\linewidth}{!}{
    \begin{tabular}{l|l}

        Summary & `\textcolor{red}{\textit{opposition leaders Ranariddh and}} 
 \\
      \infounit & \textcolor{red}{\textit{Sam Rainsy}}...wanted them abroad.' \\
       
      \hline
        Document & `\textcolor{red}{\textit{Opposition leaders}} Prince Norodom \textcolor{red}{\textit{Ranariddh}}  
 \\
        \infounit & \textcolor{red}{\textit{and Sam Rainsy}}...citing Hun Sen's threats' \\

    \end{tabular}}
     \caption{An incorrect alignment example of $\RA_{IU}$, where despite the large \textcolor{red}{\textit{word overlap}}, the two \infounit s do not mean the same. }
    \label{tab_RougeExmp}
\end{table}

\begin{table}[t]
    \resizebox{\linewidth}{!}{
    \begin{tabular}{l|l}
        Summary & `\textcolor{red}{\textit{The}} SPLC's most outstanding successes...have \\
        \infounit &  been in its civil lawsuits against hate \textcolor{red}{\textit{groups}}.' \\
        \hline
        Document & `\textcolor{red}{\textit{The}} Southern Poverty Law Center... won
 \\
       \infounit &  major legal fights against the Ku Klux Klan 
       
      \\
      & and other white supremacist \textcolor{red}{\textit{groups}}.' 
      \\

    \end{tabular}}
         \caption{A correct alignment example of \super~with small \textcolor{red}{\textit{overlap}} between the \infounit s. }
    \label{tab_supervisedExmp}
\end{table}

\section{\infounit-enhanced Sentence Alignment}
\label{sec_sentExperiments}
As mentioned in \sect \ref{sec_relatedWork}, most previous alignment approaches operate on the sentence-level.
In this section, to fairly compare to sentence-level methods,
we apply our \super~aligner to extract salient sentences, based on the predicted \infounit~alignments (denoted as ``\super$_{sent}$''). We first show, intrinsically, that our method yields a more accurate set
of salient sentences than those derived by common sentence-level ROUGE-based alignments (\sect \ref{subsec_int_sent_results}). 
Additionally, we show that the performance of a prominent alignment-based summarization model is not harmed, and even slightly improves, when trained with salient sentences extracted by our method (\sect \ref{subsec_ext_sent_results}).
Overall, this section demonstrates that our proposition-level alignments are not inferior to prior sentence-alignments, even when ported to a sentence-based setting. 

\subsection{Salient-sentence Extraction}
\label{subsec_int_sent_results}

Our method for extracting salient sentences for training a salience model first scores each document sentence based on 
the aligned \infounit s it contains. This is done by combining the alignment classification scores for these \infounit s, weighted by the respective \infounit s lengths. 
Then, we select the sentences with the highest scores, until all aligned summary \infounit s are covered by alignments of \infounit s within the selected sentences. 

We consider two characteristic baselines:

\textbf{Full summary ROUGE.} This widely used baseline, proposed by \citet{nallapati2017summarunner} and denoted here $\RA_{full}$, does not exploit explicit alignments. Rather, it greedily selects source sentences that yield the highest ROUGE with respect to the \textit{entire} reference summary, until none of the remaining candidate
sentences marginally improves the ROUGE score.

\textbf{Sent2Sent ROUGE aligner.} This baseline, denoted $\RA_{sent}$, was proposed by \citet{chen2018fastAbsSumm, Lebanoff2019ScoringSS}, yielding competitive summarization results. Similar to our ROUGE-based lexical model in \sect \ref{sec_spanExperiments}, it aligns each summary sentence to one or two document sentences with which it has the highest ROUGE score. The generated salience training data consists of all the aligned document sentences.

To intrinsically assess the quality of the salient-sentences training data generated by these methods, we compare them against a gold set of salient sentences which we derived from our manually-annotated (\infounit-level) test data. To that end, a document sentence is considered salient if it contains an aligned \infounit, as this indicates that the corresponding content is included in the summary, and hence salient.
We use two metrics to evaluate the quality of an extracted set of salient sentences, as follows. (1) $Recall$: the percent of summary \infounit s that are covered, through \infounit~alignments,
by the extracted salient sentences, which reflects the amount of covered summary information; (2) $Precision$: The percent of tokens in the extracted sentences that are part of aligned \infounit s; this reflects the proportion of salient information within the extracted sentences.
The results, in Table \ref{tab_sentSalience},
indicate that our alignment-based salient sentences match the gold set substantially better, showing higher correlation with the reference summary in terms of both recall and precision. 

\begin{table}[!tbp]
\centering
    \resizebox{\linewidth}{!}{
    \begin{tabular}{cc|c|c|c}
       & \textbf{\#sents} & R-1 & R-2 & R-L \\ \hline
      \parbox[t]{1mm}{\multirow{4}{*}{\rotatebox[origin=c]{90}{$\RA_{sent}$}}}
    & 2  & 40.06 ($\pm .22$) & 17.77 ($\pm .23$) & 35.93 ($\pm .22$) \\
    & 3  & 39.81 ($\pm .22$) & 18.05 ($\pm .21$) & 36.16 ($\pm .21$) \\
    & 4  & 37.43 ($\pm .21$) & 17.47 ($\pm .19$) & 34.36 ($\pm .20$)  \\ 
    & 5  & 34.65 ($\pm .20$) & 16.6 ($\pm .19$)  & 32.07 ($\pm .19$)  \\
    \hline

      \parbox[t]{1mm}{\multirow{4}{*}{\rotatebox[origin=c]{90}{\footnotesize{\super$_{sent}$}}}}
    & 2  & 39.76 ($\pm .23$) & 17.20 ($\pm .22$) & 35.62 ($\pm .23$) \\
    & 3  & \textbf{40.40} ($\pm .22$) & \textbf{18.26} ($\pm .22$) & \textbf{36.79} ($\pm .21$) \\
    & 4  & 38.29 ($\pm .21$) & 17.78 ($\pm .20$) & 35.23 ($\pm .20$) \\
    & 5  & 35.85 ($\pm .19$) & 17.05 ($\pm .18$) & 33.24 ($\pm .19$) \\
    \hline 
    
    \end{tabular}}
    \caption{ROUGE-1, -2 and -L results, with $\geq95\%$ confidence intervals , on CNN/DM for the $\RA_{sent}$ and \super$_{sent}$ for several predicted summary lengths.}
    \label{tab_ext_sent}
    
\end{table}

\begin{table}[!htbp]
    \resizebox{\linewidth}{!}{
    \begin{tabular}{l|c|c|c}
        \textbf{Method} & $Recall$ & $Precision$ & $F_1$  \\ \hline
         $\RA_{full}$  & 63.43 & 40.59 & 49.50  \\
         $\RA_{sent}$ & 73.88 & 36.97 & 49.27  \\
         \super$_{sent}$  (Ours)  & \textbf{75.37} & \textbf{52.03} & \textbf{61.75}  \\
    \end{tabular}}
    \caption{Salient sentence detection evaluation}
    \label{tab_sentSalience}
\end{table}

\begin{table*}[t]
    \resizebox{\linewidth}{!}{
    \begin{tabular}{l|l|l}
    \multicolumn{1}{c|}{\bfseries {Reference Summary}} & \multicolumn{1}{c|}{\bfseries {Aligned Sentence-based Summary}} & \multicolumn{1}{c}{\bfseries {Aligned Propositions-based Summary}} \\ \hline

     \makecell[l]{Tairod Nathan Webster Pugh enters\\ not guilty plea to terror-related\\ charges .}&
     \makecell[l]{Air Force veteran who allegedly tried to join\\ ISIS in Syria but was turned back by Turkish\\ authorities before he could get to \textcolor{red}{\textit{the war}}\\\textcolor{red}{\textit{-torn country entered a not guilty plea to terror}}\\\textcolor{red}{\textit{-related charges Wednesday in a federal court.}}} &
     \makecell[l]{The war-torn country entered a not\\ guilty plea to terror-related charges\\ Wednesday in a federal court.} \\ \hline
     
        \makecell[l]{Pugh flew to Turkey on January 10,\\ authorities say .} &
        \makecell[l]{\textcolor{red}{\textit{The defendant}}, a former avionics instrument\\ system specialist in the Air Force, \textcolor{red}{\textit{flew from}}\\ \textcolor{red}{\textit{Egypt to Turkey on January 10}}, weeks after\\ being fired from a job as an airplane mechanic.}  &
        \makecell[l]{The defendant flew from Egypt to\\ Turkey on January 10}\\ \hline

       \makecell[l]{Authorities allege a letter on his\\ laptop told his wife he was\\ a mujahedeen .} &
       \makecell[l]{\textcolor{red}{\textit{Among the evidence, prosecutors allege:}}\\ \textcolor{red}{\textit{Investigators discovered on his laptop}}\\ \textcolor{red}{\textit{computer a letter}} saying he wanted to `` use\\ the talents and skills given to me by Allah''\\ and a chart of points where ISIS controls.}  &
       \makecell[l]{Among the evidence prosecutors allege:\\ Investigators discovered on his laptop\\ computer a letter.}\\ \hline

    \end{tabular}}
    \caption{Illustration of alignment-based summaries, over aligned full sentences vs. propositions}
    %Selected propositions from the third column are also \textcolor{red}{\textit{highlighted}} in their original sentence in the second column.}
    \label{tab_potential_prop_summary}
\end{table*}

\subsection{Extrinsic Summarization Evaluation}
\label{subsec_ext_sent_results}

Having generated improved sentence salience data,
we wish to assess its impact on overall summarization results.
To that end, we replaced the original $\RA_{sent}$ salience training set, used within an extractive salience component in a popular competitive and highly-efficient summarization model \cite{chen2018fastAbsSumm}, with our \super$_{sent}$ training set.
Both training sets were extracted from the CNN/Daily Mail single-document summarization dataset \citep{Hermann2015TeachingMT}.

In evaluation, we generated and compared summaries at various lengths, since the number of sentences to select is commonly considered a parameter. The results, in Table \ref{tab_ext_sent}, indicate that choosing only two document sentences for the summary gives a small advantage to the ROUGE-based training data.
However, for longer summaries our \super$_{sent}$ outperforms the ROUGE-based approach in all measures, statistically significantly according to 95\% confidence intervals. In fact, the longer the summary, the larger the difference between the two aligners becomes. As the average summary length of CNN/Daily Mail is 3.8 sentences, the advantage of the \super$_{sent}$ in those lengths stresses its benefit over ROUGE. Moreover, the \super$_{sent}$ data achieved the highest global result across all summary lengths.

\section{Conclusion and Discussion}
\label{sec_conclusion}

We advocate the potential of summary-source proposition-level alignment to extract cleaner and more accurate alignments than through the common sentence-level approach.
To that end, we establish a new proposition-level alignment task by releasing high-quality dev and test datasets, and an automatically-derived training set.
Our proposed supervised baseline alignment model, trained on the released data, outperforms existing lexical and semantic similarity methods. Notably, it exhibits an excessive ability to yield abstract alignments.

These resources provide fertile ground for developing improved proposition-based alignment methods that, similar to sentence-level aligners, can supply training datasets for several summarization subtasks.
A proposition-level salience dataset, for example, can be derived by marking each aligned \infounit~in a source document as salient. Accordingly, such datasets can be used to train proposition-based models for various summarization components.

In future work, proposition-based extractive summarization has the potential to yield bullet-style summaries with optimized content (similar to CNN/DailyMail \citep{fabbri2019multinews}), albeit somewhat less coherent.
An example of such potential summary, illustrated by an oracle-system summary derived from our supervised aligner predictions on CNN/DailyMail, is shown in Table \ref{tab_potential_prop_summary}.
Alternatively, our data can contribute to the recent highlighting task \citep{Arumae2019TowardsAA, Cho2020BetterHC}, where salient information fragments are marked inside a document, thus circumventing the need to generate coherent text.
Further, propositions may be fused together to generate a coherent abstractive summary. Recently, such a cascaded approach \citep{Lebanoff2020ACA}, consisting of text fragment selection followed by a generation step, exhibited comparable or improved results over end-to-end systems.

Overall, we suggest that our released resources would encourage appealing future research on proposition-based summarization approaches, as well as on developing improved alignment models, addressing a challenging semantic matching task.

\section*{Acknowledgments}
We thank the anonymous reviewers for their constructive comments.
This work was supported in part by the German Research Foundation through the German-Israeli Project Cooperation (DIP, grant DA 1600/1-1); by the Israel Science Foundation (grant 1951/17); by a grant from the Israel Ministry of Science and Technology; and by grants from Intel Labs. MB and RP were supported by NSF-CAREER Award 1846185 and a Microsoft PhD Fellowship.

\bibliography{bibliography}

\begin{thebibliography}{27}
\expandafter\ifx\csname natexlab\endcsname\relax\def\natexlab#1{#1}\fi

\bibitem[{Arumae et~al.(2019)Arumae, Bhatia, and Liu}]{Arumae2019TowardsAA}
Kristjan Arumae, Parminder Bhatia, and Fei Liu. 2019.
\newblock Towards annotating and creating summary highlights at sub-sentence
  level.
\newblock In \emph{EMNLP}.

\bibitem[{Chen and Bansal(2018)}]{chen2018fastAbsSumm}
Yen-Chun Chen and Mohit Bansal. 2018.
\newblock \href {https://doi.org/10.18653/v1/P18-1063} {Fast abstractive
  summarization with reinforce-selected sentence rewriting}.
\newblock In \emph{Proceedings of the 56th Annual Meeting of the Association
  for Computational Linguistics (Volume 1: Long Papers)}, pages 675--686,
  Melbourne, Australia. Association for Computational Linguistics.

\bibitem[{Cho et~al.(2019)Cho, Lebanoff, Foroosh, and Liu}]{Cho2019ImprovingTS}
Sangwoo Cho, Logan Lebanoff, Hassan Foroosh, and Fei Liu. 2019.
\newblock Improving the similarity measure of determinantal point processes for
  extractive multi-document summarization.
\newblock In \emph{ACL}.

\bibitem[{Cho et~al.(2020)Cho, Song, Li, Yu, Foroosh, and
  Liu}]{Cho2020BetterHC}
Sangwoo Cho, Kaiqiang Song, Chen Li, Dong Yu, H.~Foroosh, and Fei Liu. 2020.
\newblock Better highlighting: Creating sub-sentence summary highlights.
\newblock In \emph{EMNLP}.

\bibitem[{Copeck et~al.(2006)Copeck, Inkpen, Kazantseva, Kennedy, Kipp,
  Nastase, and Szpakowicz}]{Copeck2006LeveragingD}
Terry Copeck, Diana Inkpen, Anna Kazantseva, Alistair Kennedy, Darren Kipp,
  Vivi Nastase, and Stan Szpakowicz. 2006.
\newblock Leveraging duc.
\newblock In \emph{Proceedings of the Workshop on Automatic Summarization (DUC
  2006), HLT/NAACL-2006}.

\bibitem[{Copeck et~al.(2007)Copeck, Inkpen, Kazantseva, Kennedy, Kipp, and
  Szpakowicz}]{Copeck2007CatchWY}
Terry Copeck, Diana Inkpen, Anna Kazantseva, Alistair Kennedy, Darren Kipp, and
  Stan Szpakowicz. 2007.
\newblock Catch what you can.
\newblock In \emph{Proceedings of the Workshop on Automatic Summarization (DUC
  2007), HLT/NAACL-2007}.

\bibitem[{Copeck et~al.(2008)Copeck, Kazantseva, Kennedy, Kunadze, Inkpen, and
  Szpakowicz}]{Copeck2008UpdateSU}
Terry Copeck, Anna Kazantseva, Alistair Kennedy, Alex Kunadze, Diana Inkpen,
  and Stan Szpakowicz. 2008.
\newblock Update summary update.
\newblock In \emph{Proceedings of the Workshop on Automatic Summarization (TAC
  2008)}.

\bibitem[{Copeck and Szpakowicz(2005)}]{Copeck2005LeveragingP}
Terry Copeck and Stan Szpakowicz. 2005.
\newblock Leveraging pyramids.
\newblock In \emph{Proceedings of the Workshop on Automatic Summarization (DUC
  2005), HLT/EMNLP-2005}.

\bibitem[{Fabbri et~al.(2019)Fabbri, Li, She, Li, and
  Radev}]{fabbri2019multinews}
Alexander Fabbri, Irene Li, Tianwei She, Suyi Li, and Dragomir Radev. 2019.
\newblock \href {https://doi.org/10.18653/v1/P19-1102} {Multi-news: A
  large-scale multi-document summarization dataset and abstractive hierarchical
  model}.
\newblock In \emph{Proceedings of the 57th Annual Meeting of the Association
  for Computational Linguistics}, pages 1074--1084, Florence, Italy.
  Association for Computational Linguistics.

\bibitem[{Gehrmann et~al.(2018)Gehrmann, Deng, and Rush}]{gehrmann2018bottomup}
Sebastian Gehrmann, Yuntian Deng, and Alexander Rush. 2018.
\newblock \href {https://doi.org/10.18653/v1/D18-1443} {Bottom-up abstractive
  summarization}.
\newblock In \emph{Proceedings of the 2018 Conference on Empirical Methods in
  Natural Language Processing}, pages 4098--4109, Brussels, Belgium.
  Association for Computational Linguistics.

\bibitem[{Hermann et~al.(2015)Hermann, Kocisk{\'y}, Grefenstette, Espeholt,
  Kay, Suleyman, and Blunsom}]{Hermann2015TeachingMT}
Karl~Moritz Hermann, Tom{\'a}s Kocisk{\'y}, Edward Grefenstette, Lasse
  Espeholt, Will Kay, Mustafa Suleyman, and Phil Blunsom. 2015.
\newblock Teaching machines to read and comprehend.
\newblock In \emph{NIPS}.

\bibitem[{Hirao et~al.(2018)Hirao, Kamigaito, and
  Nagata}]{hirao-etal-2018-automatic}
Tsutomu Hirao, Hidetaka Kamigaito, and Masaaki Nagata. 2018.
\newblock \href {https://doi.org/10.18653/v1/D18-1450} {Automatic pyramid
  evaluation exploiting {EDU}-based extractive reference summaries}.
\newblock In \emph{Proceedings of the 2018 Conference on Empirical Methods in
  Natural Language Processing}, pages 4177--4186, Brussels, Belgium.
  Association for Computational Linguistics.

\bibitem[{Lebanoff et~al.(2020)Lebanoff, Dernoncourt, Kim, Chang, and
  Liu}]{Lebanoff2020ACA}
Logan Lebanoff, Franck Dernoncourt, Doo~Soon Kim, W.~Chang, and Fei Liu. 2020.
\newblock A cascade approach to neural abstractive summarization with content
  selection and fusion.
\newblock In \emph{AACL/IJCNLP}.

\bibitem[{Lebanoff et~al.(2019)Lebanoff, Song, Dernoncourt, Kim, Kim, Chang,
  and Liu}]{Lebanoff2019ScoringSS}
Logan Lebanoff, Kaiqiang Song, Franck Dernoncourt, Doo~Soon Kim, Seokhwan Kim,
  Walter Chang, and Fei Liu. 2019.
\newblock Scoring sentence singletons and pairs for abstractive summarization.
\newblock In \emph{ACL}.

\bibitem[{Lin(2004)}]{lin2004rouge}
Chin-Yew Lin. 2004.
\newblock {ROUGE}: A package for automatic evaluation of summaries.
\newblock In \emph{Proc. of Workshop on Text Summarization Branches Out, Post
  Conference Workshop of ACL 2004}.

\bibitem[{Liu et~al.(2019)Liu, Ott, Goyal, Du, Joshi, Chen, Levy, Lewis,
  Zettlemoyer, and Stoyanov}]{liu2019roberta}
Yinhan Liu, Myle Ott, Naman Goyal, Jingfei Du, Mandar Joshi, Danqi Chen, Omer
  Levy, Mike Lewis, Luke Zettlemoyer, and Veselin Stoyanov. 2019.
\newblock Roberta: A robustly optimized bert pretraining approach.
\newblock \emph{arXiv preprint arXiv:1907.11692}.

\bibitem[{Nallapati et~al.(2017)Nallapati, Zhai, and
  Zhou}]{nallapati2017summarunner}
Ramesh Nallapati, Feifei Zhai, and Bowen Zhou. 2017.
\newblock Summarunner: A recurrent neural network based sequence model for
  extractive summarization of documents.
\newblock In \emph{Thirty-First AAAI Conference on Artificial Intelligence}.

\bibitem[{Nenkova and Passonneau(2004)}]{nenkova2004pyramid}
Ani Nenkova and Rebecca Passonneau. 2004.
\newblock Evaluating content selection in summarization: The pyramid method.
\newblock In \emph{Proceedings of the human language technology conference of
  the north american chapter of the association for computational linguistics:
  Hlt-naacl 2004}, pages 145--152.

\bibitem[{NIST(2014)}]{nist2014duc}
NIST. 2014.
\newblock Document understanding conferences.
\newblock \url{https://duc.nist.gov}.
\newblock Accessed: 2019-12-02.

\bibitem[{Peyrard and
  Eckle-Kohler(2017)}]{peyrard-eckle-kohler-2017-supervised}
Maxime Peyrard and Judith Eckle-Kohler. 2017.
\newblock \href {https://doi.org/10.18653/v1/P17-1100} {Supervised learning of
  automatic pyramid for optimization-based multi-document summarization}.
\newblock In \emph{Proceedings of the 55th Annual Meeting of the Association
  for Computational Linguistics (Volume 1: Long Papers)}, pages 1084--1094,
  Vancouver, Canada. Association for Computational Linguistics.

\bibitem[{Roit et~al.(2020)Roit, Klein, Stepanov, Mamou, Michael, Stanovsky,
  Zettlemoyer, and Dagan}]{roit-etal-2020-controlled}
Paul Roit, Ayal Klein, Daniela Stepanov, Jonathan Mamou, Julian Michael,
  Gabriel Stanovsky, Luke Zettlemoyer, and Ido Dagan. 2020.
\newblock \href {https://doi.org/10.18653/v1/2020.acl-main.626} {Controlled
  crowdsourcing for high-quality {QA}-{SRL} annotation}.
\newblock In \emph{Proceedings of the 58th Annual Meeting of the Association
  for Computational Linguistics}, pages 7008--7013, Online. Association for
  Computational Linguistics.

\bibitem[{Sigelman(2006)}]{sigelman2006ducview}
Sergey Sigelman. 2006.
\newblock Ducview.
\newblock
  \url{http://personal.psu.edu/rjp49/DUC2006/2006-pyramid-guidelines.html}.
\newblock Accessed: 2019-12-02.

\bibitem[{Stanovsky et~al.(2018)Stanovsky, Michael, Zettlemoyer, and
  Dagan}]{Stanovsky2018SupervisedOI}
Gabriel Stanovsky, Julian Michael, Luke Zettlemoyer, and Ido Dagan. 2018.
\newblock Supervised open information extraction.
\newblock In \emph{NAACL-HLT}.

\bibitem[{Sultan et~al.(2014)Sultan, Bethard, and Sumner}]{sultan2014back}
Md~Arafat Sultan, Steven Bethard, and Tamara Sumner. 2014.
\newblock Back to basics for monolingual alignment: Exploiting word similarity
  and contextual evidence.
\newblock \emph{Transactions of the Association for Computational Linguistics},
  2:219--230.

\bibitem[{Yang et~al.(2016)Yang, Passonneau, and de~Melo}]{Yang2016PEAKPE}
Qian Yang, Rebecca~J. Passonneau, and Gerard de~Melo. 2016.
\newblock {PEAK}: Pyramid evaluation via automated knowledge extraction.
\newblock In \emph{AAAI}.

\bibitem[{Zhang et~al.(2019)Zhang, Kishore, Wu, Weinberger, and
  Artzi}]{zhang2019bertscore}
Tianyi Zhang, Varsha Kishore, Felix Wu, Kilian~Q Weinberger, and Yoav Artzi.
  2019.
\newblock Bertscore: Evaluating text generation with bert.
\newblock \emph{arXiv preprint arXiv:1904.09675}.

\bibitem[{Zhang et~al.(2018)Zhang, Lapata, Wei, and
  Zhou}]{zhang2018neuralExtractiveSumm}
Xingxing Zhang, Mirella Lapata, Furu Wei, and Ming Zhou. 2018.
\newblock \href {https://doi.org/10.18653/v1/D18-1088} {Neural latent
  extractive document summarization}.
\newblock In \emph{Proceedings of the 2018 Conference on Empirical Methods in
  Natural Language Processing}, pages 779--784, Brussels, Belgium. Association
  for Computational Linguistics.

\end{thebibliography}
\bibliographystyle{acl_natbib}

\appendix
\label{sec:appendix}
\section{Annotation Guidelines}
\label{sec_croud_guidelines}
The crowdsourcing instructions are as follows:

\paragraph{\underline{Task 1:} Information Unit Extraction}
Your task is to divide a sentence to its standalone facts. Each fact is called ``information unit". The units must cover the whole sentence.

Each unit contains one verb and all its arguments.
In some cases, where one verb can't stand alone without another verb- the unit will contain more than one verb (e.g. ``...insisted that the ceremony take place in...").
In such cases one of the verbs uses the other as an argument.

A unit doesn't have to be grammatically valid or continuous.

For example, the sentence:
\begin{center}
\textit{Twenty-one people were injured and received treatment from MDA when an explosives-rigged car blew up Friday at Jerusalem's Mahane Yehuda market.}    
\end{center}
should be divided into:

\begin{enumerate}
\item \textit{Twenty-one people were injured}
\item \textit{Twenty-one people...received treatment from MDA}
\item \textit{an explosives-rigged car blew up Friday at Jerusalem's Mahane Yehuda market.}
\end{enumerate}

You may follow these guidelines to help you extract the information units:

\begin{enumerate}
\item Find all verbs.
\item Split the sentence according to the verbs. One verb (and all its arguments) should be included in each information unit.
\item Try to include the subject in the unit, even if the unit becomes discontinuous. It is OK that a word is used in several units, but do not repeat a whole fact twice.
\end{enumerate}

\paragraph{\underline{Task 2:} Non-Alignment Filtering}
In this task, you get one primary sentence with a bold span, and several secondary sentences.
Your task is to decide whether each one of the secondary sentences contains the bold span's significant information.
In that case, the two sentences are called "aligned".
In addition, you should mark the aligned span from the secondary sentence.
The full primary sentence is presented only for context. You only need to match the information in the bold span.

You may follow these questions to help you to decide for alignment:
\begin{itemize}
\item Does the information in the secondary sentence repeat the information in the bold span?
\item Can the bold span replace a part of the secondary sentence without changing the meaning and without deteriorating the information (except for side details)?
\item Does the secondary sentence implicate the bold span?
\end{itemize}

Notice! in order to be aligned, the sentence should talk specificaly on the same information of the bold span.

\noindent
For example, the sentence:
\begin{center}
\textbf{Prime Minister Hun Sen insisted that talks take place in Cambodia} while opposition leaders Ranariddh and Sam Rainsy, fearing arrest at home, wanted them abroad.
\end{center}
is aligned to:

\begin{center}
\ul{Cambodian leader Hun Sen on Friday rejected opposition parties' demands for talks outside the country}, accusing them of trying to ``internationalize'' the political crisis.
\end{center}
because if \textit{Hun Sen on Friday rejected opposition parties' demands for talks outside the country} it implicates \textbf{Prime Minister
Hun Sen insisted that talks take place in Cambodia}

However, the sentence:

\begin{center}
\textbf{Prospects were dim for resolution of the political crisis in Cambodia in October 1998.}
\end{center}
is not aligned to:

\begin{center}
\textit{Cambodian leader Hun Sen on Friday rejected opposition parties' demands for talks outside the country, accusing them of trying to ``internationalize'' the political crisis.}
\end{center}
although the struggle that was presented in the sentence, may point on a political crisis, if the bold span replace the aligned span we will lose significant information.
The spans should be 100\% aligned (except for side details).

\paragraph{\underline{Task 3:} Information Unit Alignment}
In this task, you get one primary sentence with a bold span, and one secondary sentence.
Your task is to highlight the maximal joint information in both sentences.

First, highlight a span from the secondary sentence that contains the bold span's significant information.
The bold span and your highlighted span are called "aligned".
Try to maximize the highlighted span as much as possible, without adding non-shared information.
The shared information must be a standalone fact (usually includes: verb, subject, object).
Names/objects only without any fact, are not considered aligned. [(`John went home';`John ate pizza') `John' is not aligned.]
However, you should add words (such as: verb, subject, object) that make the span a standalone fact, even if they are not exactly aligned to the primary sentence.

Next, highlight a maximal sub-span from the bold span (in the primary sentence) that contains only the shared information with the highlighted span from the secondary sentence.
Non-shared information should be omitted from both highlighted spans.

The full primary sentence is presented only for context. You only need to match the information in the bold span.
In case there is no joint information between the two sentences, you may choose ``Not aligned".

You may follow these questions to help you to decide for alignment:

\begin{itemize}
\item Does the information in the secondary sentence repeat the information in the bold span?
\item Can the bold span replace a part of the secondary sentence without changing the meaning and without deteriorating the information (except for side details)?
\item Does the secondary sentence implicate the bold span
\end{itemize}
Notice! in order to be aligned, the sentence should talk specifically about the same information as the bold span.

\noindent
For example, the sentence:
\begin{center}
\textbf{On Wednesday, \ul{Prime Minister Hun Sen insisted that talks take place in Cambodia}} while opposition leaders Ranariddh and Sam Rainsy, fearing arrest at home, wanted them abroad.
\end{center}
is aligned to:

\begin{center}
\ul{Cambodian leader Hun Sen} on Friday \ul{rejected opposition parties' demands for talks outside the country}, accusing them of trying to ``internationalize'' the political crisis.
\end{center}

because if \textit{Hun Sen...rejected opposition parties' demands for talks outside the country} it implicates \textit{Prime Minister
Hun Sen insisted that talks take place in Cambodia}

Please notice we omitted "On Wednesday," from the primary sentence, because it is not included in the secondary sentence.

\begin{figure*}[t!]
  \includegraphics{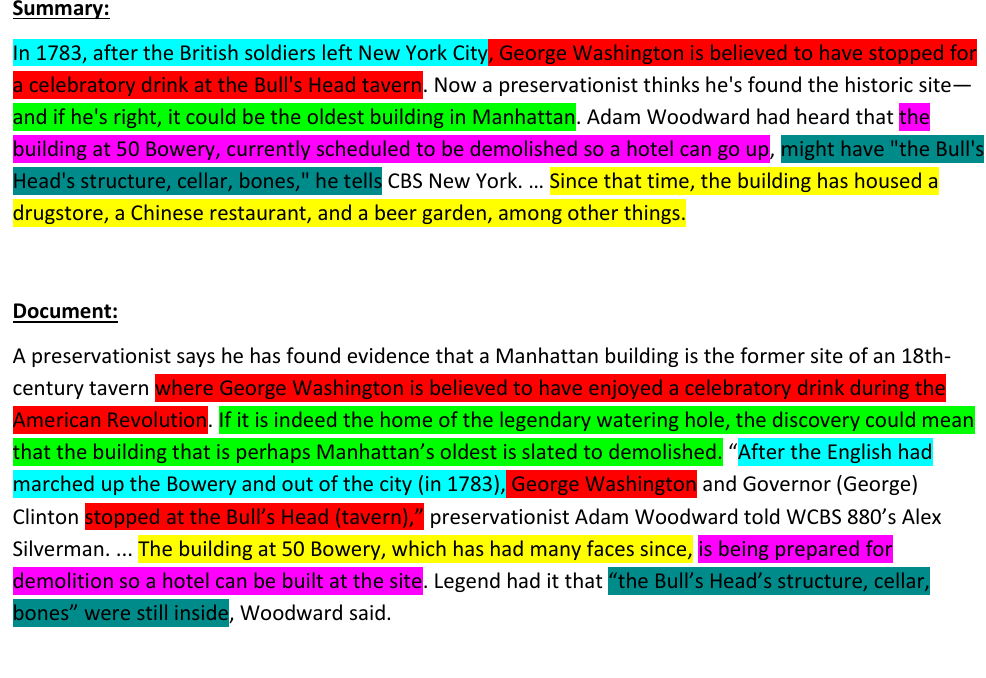}
 \caption{Example of alignments between a summary and a source document. In this example, there are no intersecting \infounit s for ease of presentation, though \infounit s can indeed intersect.}
 \label{fig_full_alignments}
\end{figure*}

\section{A Full Alignment Example}
\label{sec_full_alignment_example}
An illustration of the alignment data is presented in Figure \ref{fig_full_alignments}. Alignment pairs are marked in the same color. Although our data is pairwise only, pairs could be consolidated through a shared summary IU, which creates an alignment cluster (see red spans in the document).

\end{document}